\documentclass{article}

\usepackage{microtype}
\usepackage{graphicx}
\usepackage{subcaption}
\usepackage{enumitem}
\usepackage{booktabs} 

\usepackage{hyperref}

\usepackage[accepted]{icml2026_testing}

\usepackage{amsmath}
\usepackage{amssymb}
\usepackage{mathtools}
\usepackage{amsthm}
\usepackage{xcolor}
\usepackage{amssymb}
\usepackage{adjustbox}

\usepackage[capitalize,noabbrev]{cleveref}

\theoremstyle{plain}

\theoremstyle{definition}

\theoremstyle{remark}

\usepackage[textsize=tiny]{todonotes}

\icmltitlerunning{Computational Safety for Generative AI}

\begin{document}

\twocolumn[
  \icmltitle{Computational Safety for Generative AI: A Hypothesis Testing Perspective}



  \icmlsetsymbol{equal}{*}

  \begin{icmlauthorlist}
    \icmlauthor{Pin-Yu Chen}{comp}
  \end{icmlauthorlist}

  \icmlaffiliation{comp}{IBM Research}

  \icmlcorrespondingauthor{Pin-Yu Chen}{pin-yu.chen@ibm.com}

  \icmlkeywords{Generative AI, Computational Safety, Hypothesis Testing}

  \vskip 0.3in
]



\printAffiliationsAndNotice{}  

\begin{abstract}
AI safety is a rapidly growing area of research that seeks to prevent the harm and misuse of frontier AI technology, particularly with respect to generative AI (GenAI) tools that are capable of creating realistic and high-quality content through text prompts. Examples of such tools include large language models (LLMs) and text-to-image (T2I) diffusion models. As the performance of various leading GenAI models approaches saturation due to similar training data sources and neural network architecture designs, the development of reliable safety guardrails has become a key differentiator for responsibility and sustainability. This paper presents a formalization of the concept of \textit{computational safety}, which is a mathematical framework that enables the quantitative assessment, formulation, and study of safety challenges in GenAI through the lens of signal processing theory and methods. In particular, we explore two exemplary categories of computational safety challenges in GenAI that can be formulated as hypothesis testing problems. For the \textit{safety of model input}, we show how sensitivity analysis and loss landscape analysis can be used to detect malicious prompts with jailbreak attempts. For the \textit{safety of model output}, we elucidate how statistical signal processing can be used to detect AI-generated content. 
Finally, we discuss key open research challenges, opportunities, and the essential role of signal processing in computational AI safety.
\end{abstract}

\section{Introduction}
 Signal processing for pattern recognition and machine intelligence has played a pivotal role in ensuring the stability, security, and efficiency of numerous engineering systems and information technologies, including, but not limited to, telecommunications, information forensics and security, machine learning, data science, and control systems. With the recent advances, wide accessibility, and deep integration of generative AI (GenAI) tools into our society and technology, such as ChatGPT and the emerging agentic AI applications, understanding and mitigating the associated risks of the so-called ``frontier AI technology'' is essential to ensure a responsible and sustainable use of GenAI. In addition, as the performance of state-of-the-art GenAI models surpasses that of an average human in certain tasks, but reaches a plateau in standardized capability evaluation benchmarks due to similar training data sources and neural network architecture designs (e.g., the use of decoder-only transformers), improving and ensuring safety is becoming the new arms race among GenAI stakeholders.
In addition to ongoing activities in risk governance, policy-making, and legal regulation (e.g., EU AI Act, AI safety/security institutes, etc.), there are growing concerns about the broader socio-technical impacts  \cite{qi2024ai}. Notable examples include breaking the embedded safety alignment of GenAI models to induce misbehavior (e.g., asking large language models how to perform illegal or dangerous activities) \cite{zou2023universal} or generating inappropriate content, and misusing GenAI tools to create and spread misinformation (e.g., using text prompts to create deepfakes and influence public opinion) \cite{verdoliva2020media}. 

The main goal of this paper is to present the central message in a technically sound and rigorous manner as follows: \textit{While AI safety can be seen as a significant challenge in the context of modern GenAI technology, it is possible to reformulate many of the associated problems as a standard hypothesis testing task}. Furthermore, insights and methods developed in signal processing can be applied to improve AI safety. Specifically, we define \textit{computational AI safety} as the set of safety problems that can be formulated as hypothesis-testing tasks in signal processing.  We also provide examples of how signal processing techniques such as sensitivity analysis, subspace projection \cite{vaswani2018robust}, loss landscape  \cite{li2018visualizing}, and adversarial learning can be used to study and improve AI safety.

\begin{figure*}
    \centering
    \includegraphics[width=0.86\linewidth]{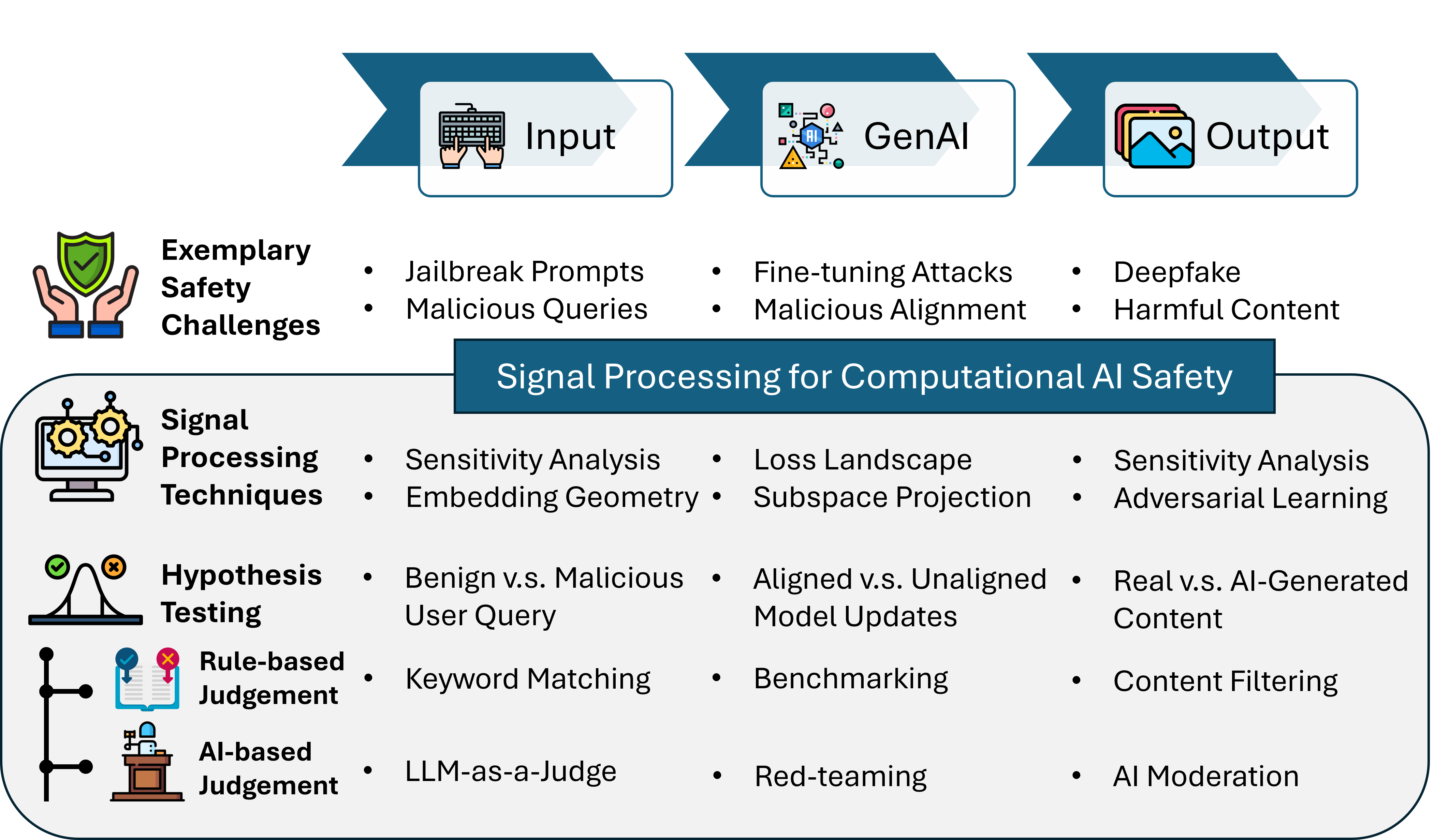}
    \caption{Overview of the framework on signal processing for computational AI safety. 
   The two safety challenges highlighted for model input and output safety in GenAI are the detection of unsafe queries and AI-generated content.  We define computational AI safety as the set of safety problems that can be formulated as hypothesis testing tasks in signal processing. We also provide examples of how signal processing techniques such as sensitivity analysis, and subspace projection can be used to improve AI safety.
   What is unique about GenAI is that the validity of the safety hypothesis (e.g., whether the input or output is safe) requires an additional judge function for certification, which can be either a rule-based approach (e.g., keyword matching) or an AI-based evaluation (e.g., LLM-as-a-judge or external contextual classifiers). The framework can be expanded with additional signal processing methods and system-level designs.}
    \label{fig:overview}
\end{figure*}

Figure \ref{fig:overview} provides an overview of our proposed signal processing framework for computational AI safety. Without loss of generality, we divide a GenAI system into three parts: the \textit{model input} that processes a user's query, the \textit{GenAI model} that performs inference and action (for agentic workflows), and the \textit{model output} produced by the model given the input. The two representative safety challenges highlighted for model input and output safety in GenAI are the detection of unsafe queries and AI-generated content.  
It is worth noting that what is unique about GenAI is that the validity of the safety hypothesis (e.g., whether the input or output is safe) requires an additional judge function for certification, which can be either a rule-based approach (e.g., keyword matching) or an AI-based evaluation (e.g., LLM-as-a-judge or external contextual classifiers). For example, when assessing the risk of jailbreak attacks, one can use refusal-related keywords to quantify the rejection rate in GenAI, or use another LLM as a judge to provide a numerical rating of the safety level of the response.

To bridge AI safety and signal processing, this paper presents an overview of computational safety for GenAI, consolidates numerous computational AI safety problems through the lens of a unified hypothesis testing framework in signal processing, and provides comprehensive technical information to elucidate how signal processing methodologies can facilitate and enhance the development of AI safety tools. This paper's signal-processing-centric perspective differs from the broad range of recent progress in attacks and defenses for GenAI in existing AI safety literature, such as \cite{chua2024ai,ma2025safety,bereska2024mechanistic}, or in-depth discussions of AI safety in particular domains \cite{sangwan2023cybersecurity,ullrich2024ai}.
The rest of the paper is organized as follows. First, we provide background on mainstream GenAI models and preliminaries on hypothesis testing and common evaluation metrics. Next, we formally introduce the signal processing framework for computational AI safety. We then examine two representative use cases in AI safety -- jailbreak detection and mitigation, and AI-generated content. Finally, we provide our perspective on the position of AI safety and discuss several ongoing research challenges and opportunities at the intersection of signal processing and AI safety.

\section{Background and Preliminaries}

\subsection{Mainstream GenAI Models}
This paper focuses on the safety challenges of two mainstream GenAI models: Large Language Models (LLMs) and Diffusion Models (DMs). LLMs are autoregressive neural networks built on self-attention-based transformers with billions of trainable parameters, which dominate text-to-text generation applications (e.g., ChatGPT) and underpin many multi-modal GenAI tools, such as vision-language models, audio-language models, and code-language models. DMs are state-of-the-art neural networks capable of 
high-quality text-to-image generation through a mathematical diffusion process, with extensions to other modalities such as video, audio, and molecular dynamics. The deployment of LLMs and DMs has two modes: direct use of the off-the-shelf GenAI model, or customized fine-tuning with some task-specific downstream datasets.
Throughout this paper, we will use $\theta$ to denote the set of trainable parameters (model weights) for GenAI models.

\paragraph*{Large language models (LLMs)}
The development of LLMs has two primary phases: pre-training and alignment. LLM pre-training refers to unsupervised machine learning on a large dataset (e.g., text extracted from the Internet) denoted by $\mathcal{D}_p$.
Let $\mathbf{x}= \{x_1,x_2,\ldots,x_T\}$ denote a sequence of tokens (i.e., ``words'' defined in the LLM vocabulary) randomly drawn from $\mathcal{D}_p$. For pre-training, the Next-token prediction (NTP) loss is commonly used, which is defined as the expected loss of the negative log-likelihood
\begin{align}
    \text{Loss}_{\text{NTP}} = - \mathbb{E}_{\mathbf{x} \sim \mathcal{D}_p}  \frac{1}{|\mathbf{x}|}\sum_{t=1}^{|\mathbf{x}|} \log p_\theta (x_t|\mathbf{x}_{<t}),
\end{align}
where $|\mathbf{x}|$ denotes the token length of $\mathbf{x}$, $\mathbf{x}_{<t} := \{x_1,x_2,\ldots,x_{t-1}\}$ denotes the tokens preceding the $t$-th token and $\mathbf{x}_{<1} := \varnothing$ (an empty set), $p_\theta (x_t| \mathbf{x}_{<t})$ is parameterized by a softmax function over all tokens.
This loss is also known as \textit{causal language modeling}, by using previous tokens to predict the next token. The family of LLMs trained with NTP loss is also called autoregressive LLMs. After pre-training on $\theta$, an LLM undergoes the alignment process to teach the model to follow instructions and value alignment. Safety alignment is also implemented in this stage to guide the model in refusing to respond to unsafe queries. Specifically, for supervised fine-tuning (SFT) (or instruction-tuning),
let $(\mathbf{x},\mathbf{y})$ denote a pair of model input and the desired model output randomly drawn from a dataset $\mathcal{D}_a$, the SFT loss is used to update the pre-trained model weights by generalizing the NTP loss as
\begin{align}
    \text{Loss}_{\text{SFT}} = - \mathbb{E}_{(\mathbf{x},\mathbf{y}) \sim \mathcal{D}_a}  \frac{1}{|\mathbf{y}|}\sum_{t=1}^{|\mathbf{y}|} \log p_\theta (y_t|\mathbf{x},\mathbf{y}_{<t}).
\end{align}
Here, $\mathbf{x}$ can also be conceived as the \textit{context}, which is used to model the conditional probability of generating 
$\mathbf{y}$ given $\mathbf{x}$  as
$p_\theta (\mathbf{y} | \mathbf{x}) = \Pi_{t=1}^{|\mathbf{y}|}  p_\theta (y_t|\mathbf{x}, \mathbf{y}_{<t})$. In addition to SFT, other forms of alignment include strategies such as reinforcement learning with human feedback (RLHF) \cite{ouyang2022training} and direct preference optimization (DPO) \cite{rafailov2024direct}.

\paragraph*{Diffusion Models (DMs)}
During training, DMs follow a mathematical diffusion process to gradually encode data samples into random Gaussian noises, and train a neural network to learn to decode (reconstruct) the data samples based on multiple denoising steps. Furthermore, DMs can be integrated with text-based prompts to guide the generation process.
We briefly explain the basics of DM as follows. Let $t$ denote the diffusion step and let
$\mathbf{x}_t$ be the data sample (or its latent representation) obtained via a forward diffusion process subject to noise injection.
Given a text prompt (context) $c$, we denote $\mathbf{\epsilon}_\theta(\mathbf{x}_t|c)$ as the noise generator parameterized by a neural network $\theta$. A reverse diffusion process is also used to estimate the underlying noise. In its simplest form, the diffusion process is given by
\begin{align}
\hat{\mathbf{\epsilon}}_\theta(\mathbf{x}_t|c) = (1-\lambda) \cdot  {\epsilon}_{\theta}(\mathbf x_t | \emptyset ) + \lambda \cdot {\epsilon}_{\theta}(\mathbf x_t | c ),
    \label{eq: condition_diffusion}
\end{align}
where $\hat{\mathbf{\epsilon}}_\theta(\mathbf{x}_t|c)$ signifies the ultimate noise estimation attained by utilizing the DM conditioned on $c$, $ \lambda \in [0,1]$ is a guidance weight, and ${\epsilon}_\theta(\mathbf x_t | \emptyset ) $ represents the contribution of the unconditional DM.
During inference, the DM initializes with a standard Gaussian noise $z_T \sim \mathcal{N}(\mathbf{0}, \mathbf{I})$ with zero mean and identity covariance matrix $\mathbf{I}$. The noise $z_T$ is then  denoised using $\hat{\epsilon}_{\theta}(\mathbf x_T | c)$ to obtain $z_{T-1}$. This procedure is repeated to 
generate the authentic data at $t = 0$. A commonly used training objective for DM is the mean-squared-error (MSE), defined as 
\begin{align}
\text{Loss}_{\text{MSE}}  =  \mathbb{E}_{t, \epsilon \sim \mathcal{N}(\mathbf{0}, \mathbf{I})}[\| \epsilon - \epsilon_{\boldsymbol \theta}(\mathbf x_t | c) \|_2^2].
\label{eq: diffusion_loss}
\end{align}
For simplicity, we omit the expectation over the training data $\{(\mathbf{x},c) \}$ for DM training.

\begin{table*}[]
\caption{Examples of AI safety problems that can be formulated as binary hypothesis testing tasks.}
\centering
\begin{adjustbox}{max width=0.99\textwidth}
\begin{tabular}{@{}l|l|l@{}}
\toprule
Problem Domain       & Alternative Hypothesis ($\mathcal{H}_1$)        & Null Hypothesis ($\mathcal{H}_0$)              \\ \midrule
Jailbreak            & Model input attempts to bypass safety guardrails & Legitimate  model input                             \\
AI-generated Content & AI-generated sample                             & Real (not AI-generated) sample                 \\
Model Fine-tuning    & Model updates compromise safety alignment      & Model updates are legitimate                   \\
Watermark            & Data sample is watermarked                      & Data sample is not watermarked                 \\
Membership Inference & A data sample has been used in model training  & A data sample has not been used in model training  \\
Data Contamination   & A dataset has been used in model training      & A dataset has not been used in model training \\ \bottomrule
\end{tabular}
\end{adjustbox}
\label{table:HT}
\end{table*}

\subsection{Hypothesis Testing and Common Evaluation Metrics}

\paragraph*{Generative Hypothesis Testing}
Hypothesis testing refers to statistical methods for data-driven validation among a set of mutually exclusive hypotheses, of which only one is true. It is a classic methodology that is widely used in signal processing, such as detection and estimation problems. 
In particular, binary hypothesis testing involves an alternative hypothesis $\mathcal{H}_1$ and a null hypothesis $\mathcal{H}_0$. In principle, a test statistic $s$ is derived from the observed data samples and is compared to a threshold $\eta$ for testing $\mathcal{H}_1$ versus $\mathcal{H}_0$. 
Unlike classic signal processing problems, what is unique about computational safety for GenAI is that a hypothesis of interest is context-dependent and usually cannot be explicitly defined in clear mathematical forms.
For example, does the user query (model input) violate the usage policy, or
does the model output contain any harmful content?
To address this issue, it is common practice to employ a judge function $J(\mathbf{x},\mathbf{y}) \in \{0,1\}$ for hypothesis validation, which takes a pair of model input and output $(\mathbf{x},\mathbf{y})$ and returns a binary label that serves as a proxy for the ground-truth hypothesis. The judge function $J$ can be as simple as a rule-based system, such as flagging problematic context or content using keyword filtering. It can also be a complex and opaque decision-making process performed by another AI model, such as LLM-as-a-judge. However, these judge functions have their limitations. Rule-based judge functions often fail to generalize in sophisticated situations that require deep contextual understanding or reliability against intentional manipulation, while AI-based solutions may inherit data biases from the training data and exhibit unfairness or hallucination in their judgments \cite{ye2024justice}.
We refer to the scenario of using an AI-based judge function as a 
\textit{generative hypothesis testing} problem setting. Verifying and improving the accuracy of the judge functions is an important topic in computational AI safety,  but it is also beyond the scope of this paper.

\paragraph*{Common Evaluation Metrics}
Let $\hat{\mathcal{H}}$ denote the predicted hypothesis using a test statistic $s$ and let $\mathcal{H}$ be the ``actual'' hypothesis determined by a judge function. Let ``1'' denote True and ``0'' denote False in binary hypothesis testing.
Given an instance, the test result is called a true positive (TP) if $\hat{\mathcal{H}}=\mathcal{H}=1$, a true negative (TN) if $\hat{\mathcal{H}}=\mathcal{H}=0$, a false positive (FP) if $\hat{\mathcal{H}}=1$ but $\mathcal{H}=0$, and a false negative (FN) if $\hat{\mathcal{H}}=0$ but $\mathcal{H}=1$. Extending the evaluation to a set of instances using the same test statistic, and denoting the number of instances of a test result by $\#$.
The true positive rate (TPR) is defined as $\textsf{TPR}=\frac{\# \text{TP}}{\# \text{TP} + \# \text{FN} }$, the false positive rate is $\textsf{FPR}=\frac{\# \text{FP}}{\# \text{FP} + \# \text{TN} }$, the false negative rate is $\textsf{FNR}=\frac{\# \text{FN}}{\# \text{FN} + \# \text{TP} }$, and the true negative rate is $\textsf{TNR}=\frac{\# \text{TN}}{\# \text{TN} + \# \text{FP} }$. Accuracy 
measures the overall correctness of the predictions, which is defined as the ratio of the correctly tested instances to the total number of instances, $\textsf{accuracy}=\frac{\# \text{TP} + \# \text{TN}}{\# \text{TP} + \# \text{TN} + \# \text{FP} + \# \text{FN}}$. 
The equal error rate (EER) is defined as the point
on the receiver operating characteristic (ROC) curve where FPR $=$ FNR, which is equivalent to $\textsf{EER} = \frac{\textsf{FPR}+\textsf{FNR}}{2}$. Furthermore, by varying the threshold $\eta$ for the test statistic $s$ in hypothesis testing, one can obtain the area under the receiver operating characteristic curve (AUROC), which provides a threshold-independent evaluation of the capability in hypothesis testing and reflects the trade-off between TPR and FPR. Another popular evaluation metric is to report TPR@(FPR$=\alpha\%$) or FPR@(TPR$=\beta \%$) for some $\alpha$ and $\beta$ by setting the threshold to meet the criterion.

\section{Hypothesis Testing for Computational AI Safety}

With the necessary background and preliminaries introduced in the previous section, we formally present a set of signal processing techniques that can be applied to evaluate and improve computational AI safety. Specifically, Table \ref{table:HT} shows a non-exhaustive list of problems in AI safety that can be formulated as a binary hypothesis testing task in a unified way. Jailbreak is concerned with detecting model inputs that attempt to bypass safety guardrails. AI-generated Content tests whether a data sample is AI-generated. Model Fine-tuning checks if model updates would compromise safety alignment. Watermark verifies if a data sample is watermarked. Membership Inference examines if a data sample has been used for model training, which is highly relevant for machine unlearning. Data Contamination looks into whether a data set has been used for model training, which is critical for ensuring the authenticity of model evaluations on public benchmarks. This paper focuses on the first three problem areas in Table \ref{table:HT}. We then present a set of signal processing techniques that can be used to study computational AI safety.

\paragraph*{Sensitivity Analysis} Inspired by the well-known principle of studying the robustness of signal processing methods in the presence of noise,
 sensitivity analysis quantifies the amount of change in data representations subject to data manipulation. Let $\mathbf{x}$ be a data sample and let $g(\mathbf{x})$ denote the vectorized continuous latent representation (i.e., embedding) of $\mathbf{x}$ obtained from a deep learning model $g$. 
 There are two common approaches to measuring sensitivity. The first one is to use a metric $M(g(\mathbf{x}),g(\mathcal{T}(\mathbf{x})))$ to compare the similarity (or difference) between a pair of data embedding $g(\mathbf{x})$ and its manipulated version $g(\mathcal{T}(\mathbf{x}))$, where the function $\mathcal{T}$ denotes a transformation operation on $\mathbf{x}$. In particular, additive Gaussian noise perturbation is a common approach to implement $\mathcal{T}$.
 The second approach is to compute the gradient of a loss function $\ell$ evaluated at $g(\mathbf{x})$ (wherever applicable), denoted by $\nabla \ell(g(\mathbf{x}))$, and use the
 gradient norm $\| \nabla \ell(g(\mathbf{x})) \|$ as a measure of sensitivity. For AI-generated image detection, Ricker et al. implement $\mathcal{T}$ as an image reconstruction function based on an autoencoder and assign $M$ as the reconstruction error for the test statistic \cite{ricker2024aeroblade}. 
 He et al. use an image foundation model for feature extraction and implement additive Gaussian noise for sensitivity analysis to detect AI-generated images \cite{he2024rigid}, where the metric $M$ is the cosine similarity.
 For jailbreak, Hu et al. compute the gradient norm of an affirmation loss on each token of the model input to identify and mitigate jailbreak prompts \cite{hu2024token}.


\paragraph*{Subspace Modeling} Subspace modeling involves learning and processing complex (often high-dimensional) data in a compressed low-dimensional subspace. Classic examples include principal component analysis (PCA), sparse signal recovery, subspace tracking, information geometry, and manifold learning, to name a few.
In \cite{hsu2024safe}, Hsu et al. leverage the technique of subspace projection to mitigate safety degradation during model fine-tuning. Given a pair of base and aligned LLM, denoted as $\{\theta_b,\theta_a\}$, where they share the same neural network architecture, and $\theta_b$ is the model obtained after the pre-training phase, whereas $\theta_a$ is obtained after fine-tuning  $\theta_b$ with alignment data. Such a pair of model weights is often publicly available on open-weight model releases. They model the subspace for alignment as the span of the difference between $\theta_b$ and $\theta_a$, formally defined as $V = \theta_a - \theta_b$. 
The weight difference $V$ encapsulates the effect of effort that leads a pre-trained model to become an aligned model.  Next, 
when the aligned model $\theta_a$ is to be fine-tuned for a downstream task, incurring a weight change denoted by $\Delta \theta$. Before $\Delta \theta$ is applied to $\theta_a$, the similarity between $V$ and $\Delta \theta$ is checked for each layer in the neural network. If the similarity is above a threshold, indicating that the model update $\Delta \theta$ is aligned with the subspace spanned by $V$, the model update is performed as is. Otherwise, the model update is performed by first projecting $\Delta \theta$ onto the subspace spanned by $V$ and then adding the projected update to the aligned model $\theta_a$. By doing so, when compared to naive model fine-tuning without subspace projection, possible safety degradation can be mitigated while maintaining high capability on the downstream task.

\paragraph*{Loss Landscape Analysis} Given a loss function $f: \mathbb{R}^d \mapsto \mathbb{R}$, loss landscape analysis refers to the exploration of the loss function by perturbing the input of $f$ with a set of random and orthogonal directions. For the purpose of loss landscape visualization, the number of perturbations is often limited to 1 or 2 dimensions. The exploration of loss landscape for deep neural networks has facilitated the understanding and advancement of generalization errors, optimization trajectories, and model merging and ensembling
\cite{li2018visualizing,keskar2016large}. Consider the case of a two-dimensional loss landscape visualization centered on a point $\mathbf{u} \in \mathbb{R}^d$ as an input to $f$. The loss landscape $F$ is obtained by perturbing $\mathbf{u}$ along two random orthogonal unit vectors $\mathbf{v}_1$ and $\mathbf{v}_2$, where $\alpha$ and $\beta$ represent the respective coefficients and evaluating them with $f$, which is defined as
\begin{align}
    F(\alpha,\beta|\mathbf{u}) = f(\mathbf{u} + \alpha \cdot \mathbf{v}_1 + \beta \cdot \mathbf{v}_2 ).
\end{align}
For jailbreak prompt detection, Hu et al. \cite{hu2024gradient} define and explore the loss landscape in the space of model input embeddings for eliciting LLM refusals to distinguish benign v.s. malicious user queries. For model fine-tuning, Peng et al. \cite{peng2024navigating}  use loss landscape analysis to uncover the existence of a ``safety basin'' in the weight space of an aligned LLM, a bowl-shaped region with a flat loss landscape as measured by jailbreak benchmarks. Such a landscape is unique to safety, as the landscape of capability 
is drastically different from the shape of a bowl.


\paragraph*{Adversarial Learning} In the context of safety, adversarial learning refers to the methodology of introducing a virtual adversary to explore and mitigate potential risks of an AI system throughout the development and deployment lifecycle \cite{chen2023book}, including activities such as creating evaluation benchmarks, identifying vulnerabilities, active testing (dynamic red-teaming), model patching, and system hardening. For example, the virtual adversary can be another AI model, an optimization-based algorithm, or a training environment that provides safety assessment as a reward function. In particular, adversarial robustness measures the worst-case performance of a target AI system in the presence of a virtual adversary, where the interplay between the adversary and the defender (the target AI system to be protected) can be formulated as a two-player game, and the learning of the optimal strategy for each player can be studied through the lens of robust optimization and statistics. To search for effective jailbreak prompts against LLMs, Zou et al. \cite{zou2023universal} formulate this task as an optimization problem that finds the best string of tokens to be inserted as a suffix to user queries. Chao et al. \cite{chao2023jailbreaking} use an LLM as the attacker and another LLM as the judge to guide the process of finding successful jailbreak prompts against a target LLM. Xiong et al. \cite{xiong2024defensive} propose a jailbreak mitigation strategy by adding a text-based defensive prompt patch after the user query, where the design of such a patch is formulated and solved by casting it as a robust (min-max) optimization problem. For AI-generated text detection, Hu et al. \cite{hu2023radar} train a robust detector by imposing a paraphraser as a virtual adversary.

\begin{figure*}
    \centering
    \includegraphics[width=1\linewidth]{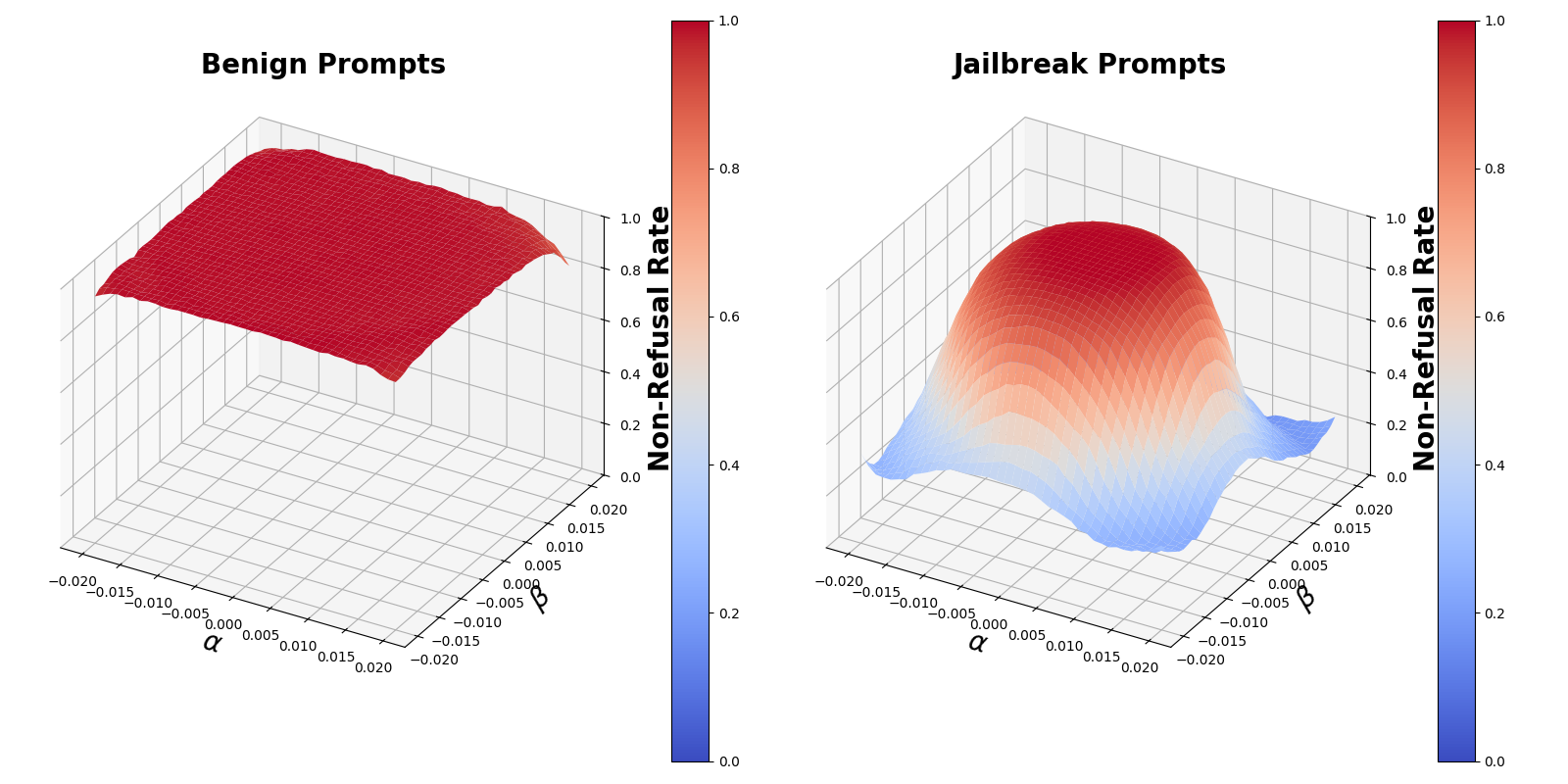}
    \caption{Loss landscape analysis for benign and jailbreak prompts. We use the token embeddings of Vicuna-7B-V1.5 to compute the non-refusal rate of model responses generated from perturbed input embeddings, by interpolating two random directions with additive Gaussian noise in the token embedding space, where the perturbation strengths are denoted by $\alpha$ and $\beta$. The results are averaged over 100 prompts. The benign prompts are sampled from Chatbot Arena, and the jailbreak prompts are generated by the greedy coordinate gradient (GCG) attack \cite{zou2023universal}. The analysis shows that the jailbreak prompts are more sensitive to Gaussian perturbations than the benign prompts.}
    \label{fig:gc}
\end{figure*}

\section{Model Input Safety: Jailbreak Prompt Detection and Mitigation}

In general, a jailbreak prompt is a user query (model input) designed to bypass the embedded safety guardrail of a GenAI system, where the system can be as simple as an open-weight LLM, or an integrated closed-source LLM with additional moderation mechanisms in place. Examples of jailbreak attack strategies include role-playing, changing system prompts, adding distractive context, and instructing another LLM to design and generate prompts. A judge function inspects the output of the target GenAI system to evaluate the success of a jailbreak prompt. A rule-based judge function uses keyword matching and declares the attack in vain if the output contains some keywords like ``Sorry'', ``I can't fulfill'', etc. However, such a judgment lacks a comprehensive understanding of the context and might overlook jailbreaks that force the target model to respond in an apologetic tone. On the other hand, one can use another LLM or a customized LLM (i.e., an LLM guard model) as a judge to evaluate whether the output is safe or unsafe. While LLM-based judge functions possess contextual awareness, they still lack transparency about the decision and may have their own biases.  Nevertheless, despite their limitations, these judge functions are poised as hypothesis-testing verifiers in our computational AI safety framework. 

\paragraph*{Safety-Capability Trade-off}
When defending against jailbreak attempts, an implicit requirement of a practical mitigation approach is the ability to effectively identify and thwart jailbreak prompts while minimizing the impact on the capability of the protected model. A model that is over-aligned to cause over-rejection of benign requests (i.e., high safety but low capability), or performant but under-aligned to be easily jailbroken (i.e., high capability but low safety), is undesirable. Therefore, exploring and optimizing the safety-capability trade-off of a deployed jailbreak mitigation is essential to computational AI safety. Next, we present two methods for detecting and mitigating jailbreak prompts to achieve a good safety-capability tradeoff.

\paragraph*{Detecting Jailbreak Prompts via Loss Landscape Analysis}
In \cite{hu2024gradient}, Hu et al. apply loss landscape analysis to the embedding of jailbreak prompts and the responses of an LLM. They show that jailbreak prompts exhibit different characteristics than benign (safe) prompts in the landscape studied, and this finding can be used to detect jailbreak prompts. Specifically, given a prompt $\mathbf{x}$ of token length $|\mathbf{x}|$, an LLM first embeds the tokens of $\mathbf{x}$ into a token embedding $\mathbf{E}(\mathbf{x}) \in \mathbb{R}^{|\mathbf{x}| \times d}$, where $d$ is the embedding dimension. The function $g$ denotes an LLM, and $g(\mathbf{E}(\mathbf{x}))$ means the response of $g$ given the context $\mathbf{x}$. A judge function $J(g(\mathbf{E}(\mathbf{x})),\mathbf{x}) \in \{0,1\}$ is used for safety verification, where ``$1$'' means $\mathbf{x}$ is a successful jailbreak prompt that causes harmful behavior of $g$, and ``$0$'' otherwise. Finally, Hu et al. explore the loss landscape around $J(g(\mathbf{E}(\mathbf{x})),\mathbf{x})$ by defining a non-refusal loss on $\mathbf{x}$ as $f(\mathbf{x}|g, J) = 1 - \frac{1}{N} \sum_{i=1}^N J(g^{(i)}(\mathbf{E}(\mathbf{x})),\mathbf{x})$, which signifies the average non-refusal rate over $N$ independent inferences of $g$ on $\mathbf{x}$ (each inference is indexed by $i$), taking into account the randomness in the decoding process (e.g., probabilistic sampling of the next token). Finally, the two-dimensional loss landscape function centered on the input embedding $\mathbf{E}(\mathbf{x})$ is computed by $F(\alpha,\beta|\mathbf{E}(\mathbf{x})) = f(\mathbf{E}(\mathbf{x}) \oplus \alpha \cdot \mathbf{v}_1 \oplus \beta \cdot \mathbf{v}_2 |g,J)$, where $\mathbf{v}_1, \mathbf{v}_2 \in \mathbb{R}^d$ are two random directions sampled from a standard multivariate Gaussian distribution, and the notation $\oplus$ denotes the row-wise broadcast function that adds a vector to each row of $\mathbf{E}(\mathbf{x})$.

Figure \ref{fig:gc} shows the average loss landscapes of benign and jailbreak prompts evaluated with Vicuna-7B-V1.5 \cite{zheng2024judging}, a fine-tuned LLM based on Meta's Llama-2-7B model \cite{touvron2023llama}. The benign prompts are 100 queries sampled from Chatbot Arena\footnote{https://lmarena.ai/}, while the 100 jailbreak prompts are generated by the Greedy Coordinate Gradient (GCG) attack on AdvBench for suffix optimization \cite{zou2023universal}. Keyword matching is used as the judge function \cite{hu2024gradient}. The results show that benign and malicious prompts differ significantly in their respective loss landscapes. The loss landscape of benign prompts is flat and close to 1, suggesting that the perturbed input embeddings are unlikely to cause rejections. On the other hand, malicious prompts are sensitive to embedding perturbations. The average non-refusal rate peaks at the origin (because most jailbreak attempts are successful), but quickly declines as the input embeddings are perturbed away from the origin, suggesting that the perturbed embeddings of malicious prompts are more likely to be refused by the LLM.
Based on the loss landscape analysis, Hu et al. propose a jailbreak prompt detection method called Gradient Cuff \cite{hu2024gradient}. The detection of Gradient Cuff consists of two steps. In the first step, the detector flags a jailbreak if the non-refusal loss $f$ is less than $0.5$, which checks the tendency of the LLM to reject a query. In the second step, the detector computes the gradient norm of $f$ at the input embedding $\mathbf{E}(\mathbf{x})$ and uses it to quantify the changes in the loss landscape under random perturbations.  If the gradient norm is greater than a certain threshold, the query is predicted to be a jailbreak prompt. In practice, such a threshold can be determined by controlling the FPR evaluated on a validation dataset of benign prompts. In addition, if the judge function $J$ is not differentiable, such as in the case of keyword matching, Gradient Cuff uses zeroth-order optimization \cite{liu2020primer}, another signal processing technique for gradient estimation using only function evaluations, to approximate the gradient norm.

\begin{figure*}[t]
\centering
  \subfloat[Safety-capability analysis]{%
       \includegraphics[width=0.54\linewidth]{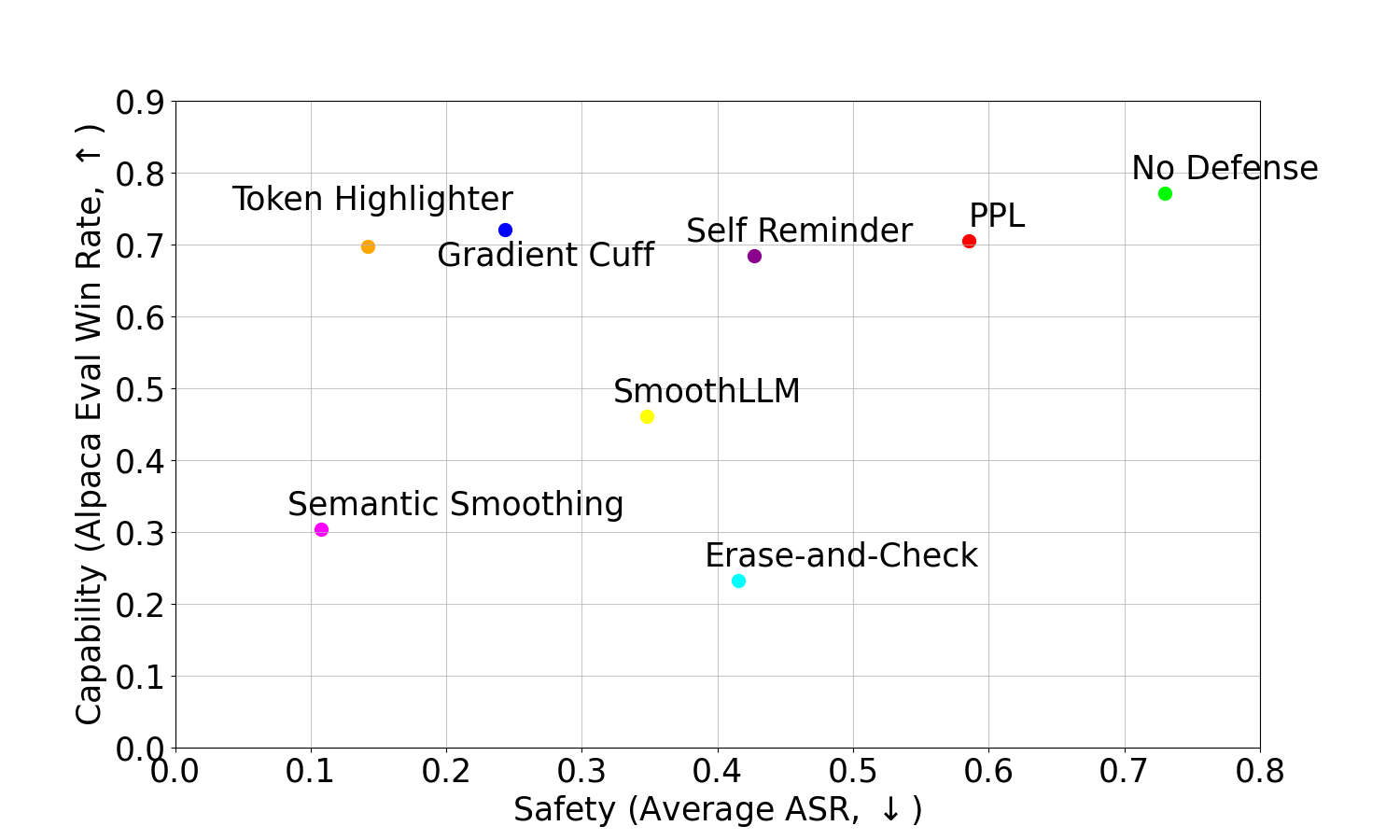}}
         \subfloat[Run time analysis]{%
       \includegraphics[width=0.45\linewidth]{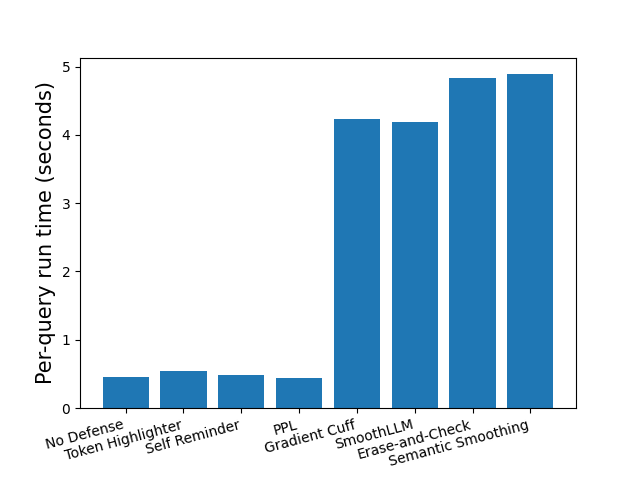}}
    \caption{Comparison of jailbreak prompt detection and mitigation methods. (a) Safety-capability trade-offs. The safety performance is evaluated by the attack success rate (ASR) averaged over 6 jailbreak attacks, and the capability performance is evaluated by the win rate in Alpaca Eval \cite{alpaca_eval}. A higher win rate and lower ASR means a better approach. See the ``Performance Evaluation'' paragraph for details). (b) Per-query run time analysis (seconds). Overall, Token Highlighter is the most economical method that best balances the safety-capability trade-off with lightweight compuation cost.}
    \label{fig:TH}
\end{figure*}

\paragraph*{Mitigating Problematic Tokens via Sensitivity Analysis}
Beyond jailbreak prompt detection, Hu et al. explore a jailbreak mitigation framework called Token Highlighter \cite{hu2024token} to identify and attenuate problematic tokens that could lead to jailbreaks. In particular, based on the principle that successful jailbreak prompts trick the target LLM into starting with an affirmative answer like ``Sure, here’s...'' \cite{zou2023universal}, Token Highlighter defines an affirmation loss as $\text{Loss}_{\text{Aff}}=-\log p_\theta(\mathbf{y}_a|
\mathbf{x})$, where $\mathbf{x}$ is the context (query), $\mathbf{y}_a$ is an affirmation phrase, and $\log p_\theta(\cdot)$ is the log-likelihood computed by an LLM parameterized by $\theta$. The phrase $\mathbf{y}_a =$ ``Sure, I’d like to help you with this.'' is the default choice in \cite{hu2024token}. With the affirmation loss $\text{Loss}_{\text{Aff}}$, the influence of each token in $\mathbf{x}$
is measured by taking the gradient of $\text{Loss}_{\text{Aff}}$ with respect to the input embedding matrix $\mathbf{E}(\mathbf{x}):= [\mathbf{e}_1; \mathbf{e}_2; \cdots \mathbf{e}_{|\mathbf{x}|}]^T$ and computing the gradient norm of each token, formally defined as $\{\| \nabla_{\mathbf{e}_j}  \text{Loss}_{\text{Aff}}\| \}_{j=1}^{|\mathbf{x}|}$, where $\mathbf{e}_j \in \mathbb{R}^d$ is the token embedding of the $j$-th token in $\mathbf{x}$. Next, Token Highlighter takes the top $Q$\% of tokens in $\mathbf{x}$ and proposes the concept of \textit{soft removal} by shrinking the embeddings of these most influential tokens by $\gamma$, where $\gamma \in [0,1)$. Finally, the LLM returns the response of the attenuated input embedding after soft removal as the ultimate model output. In its evaluation, the soft removal effectively mitigates jailbreak attempts for malicious queries while balancing the generation quality for benign queries.

\paragraph*{Performance Evaluation}
We choose Vicuna-7B-V1.5 \cite{zheng2024judging} as the target LLM to study the safety-capability trade-offs of different jailbreak prompt mitigation methods. Vicuna-7B-V1.5 is an open-weight LLM with 7 billion parameters, with good capability and mediocre safety alignment, making it an ideal candidate for studying jailbreak attack and defense strategies.  The following 6 jailbreak attacks are used for evaluation:
\begin{itemize}[leftmargin=*]
    \item GCG \cite{zou2023universal} trains and appends an adversarial suffix to a query using a greedy coordinate gradient (GCG) algorithm.
    \item AutoDAN \cite{liu2024autodan} designs a hierarchical genetic algorithm for jailbreak prompt optimization.
    \item PAIR \cite{chao2023jailbreaking} uses LLMs as an attacker and a judge for
    prompt automatic iterative refinement (PAIR).
    \item TAP \cite{mehrotra2025tree} extends PAIR with reasoning ability based on Tree of Attacks with Pruning (TAP). 
    \item AIM\footnote{The AIM prompt is obtained from https://github.com/alexalbertt/jailbreakchat} is a role-playing jailbreak attack by asking the target LLM to be Always Intelligent and Machiavellian (AIM) to give an unfiltered response to any request.

    \item Many-shot \cite{anil2024many} uses in-context examples for jailbreaking, where multiple fake dialogues between a user and an AI assistant (with affirmative responses) are provided as prefixes to the actual user query.
\end{itemize}
We also consider the following 8 mitigation strategies in the evaluation.
\begin{itemize}[leftmargin=*]
    \item No Defense means the original LLM without any mitigation strategy.
    \item PPL \cite{jain2023baseline} uses a perplexity (PPL) filter on the input query for detection.
    \item Erase-and-Check \cite{kumar2024certifying} independently erases some tokens of the original query multiple times and checks if any of the remaining queries would be rejected by the target LLM.
    \item SmoothLLM \cite{robey2023smoothllm} randomly perturbs multiple copies of the original query and then aggregates the corresponding predictions for detection.
    \item Semantic Smoothing \cite{ji2024defending} extends SmoothLLM with multiple semantically transformed copies. 
    \item Self Reminder \cite{xie2023defending} changes the system prompt to remind the target LLM to respond responsibly.
    \item Gradient Cuff \cite{hu2024gradient} explores the refusal loss landscape for detection.
    \item Token highlighter \cite{hu2024token} uses the gradient of an affirmation loss and soft removal for mitigation.
\end{itemize}
For capability assessment, we report the \textit{Win Rate} measured on Alpaca Eval \cite{alpaca_eval}, where the protected LLM's response is compared with that of OpenAI's Text-Davinci-003 model, and the quality is judged by GPT-4. For safety evaluation, we sampled 100 harmful queries from AdvBench \cite{zou2023universal} as the prototype prompts to implement jailbreak attacks. We report the \textit{Attack Success Rate} (ASR) of the input queries and their responses averaged over the aforementioned 6 jailbreak attacks, where the LLaMA-Guard-2 model \cite{inan2023llama} judges the success. An ideal mitigation strategy would have low ASR and high capability. We use the same implementations for the attacks and defenses as in \cite{hu2024token}.

Figure \ref{fig:TH}(a) shows the safety-capability trade-offs of the 8 mitigating strategies. Compared to the original (No Defense) LLM, it can be observed that Gradient Cuff and Token Highlighter strike a better trade-off between the win rate and ASR than other methods, achieving a significant reduction in the ASR while maintaining a similar level of capability. On the other hand, other methods, such as semantic smoothing, can further reduce the ASR at the cost of a more than 50\% drop in the win rate, which could severely compromise the performance of the protected LLM.
Figure \ref{fig:TH}(b) compares the per-query run time of each mitigation method. 
Mitigation strategies such as Gradient Cuff, SmoothLLM, Semantic Smoothing, and Erase-and-Check require additional inference on multiple modified copies of the original query, incurring high computation cost and latency. PPL and Self Reminder are cost-effective, but their defense performance is less effective than others. Overall, Token Highlighter is the most economical mitigation strategy, as it achieves a good tradeoff between safety and capability, and has lightweight computation.

\begin{figure*}[t]
    \centering
    \includegraphics[width=1\linewidth]{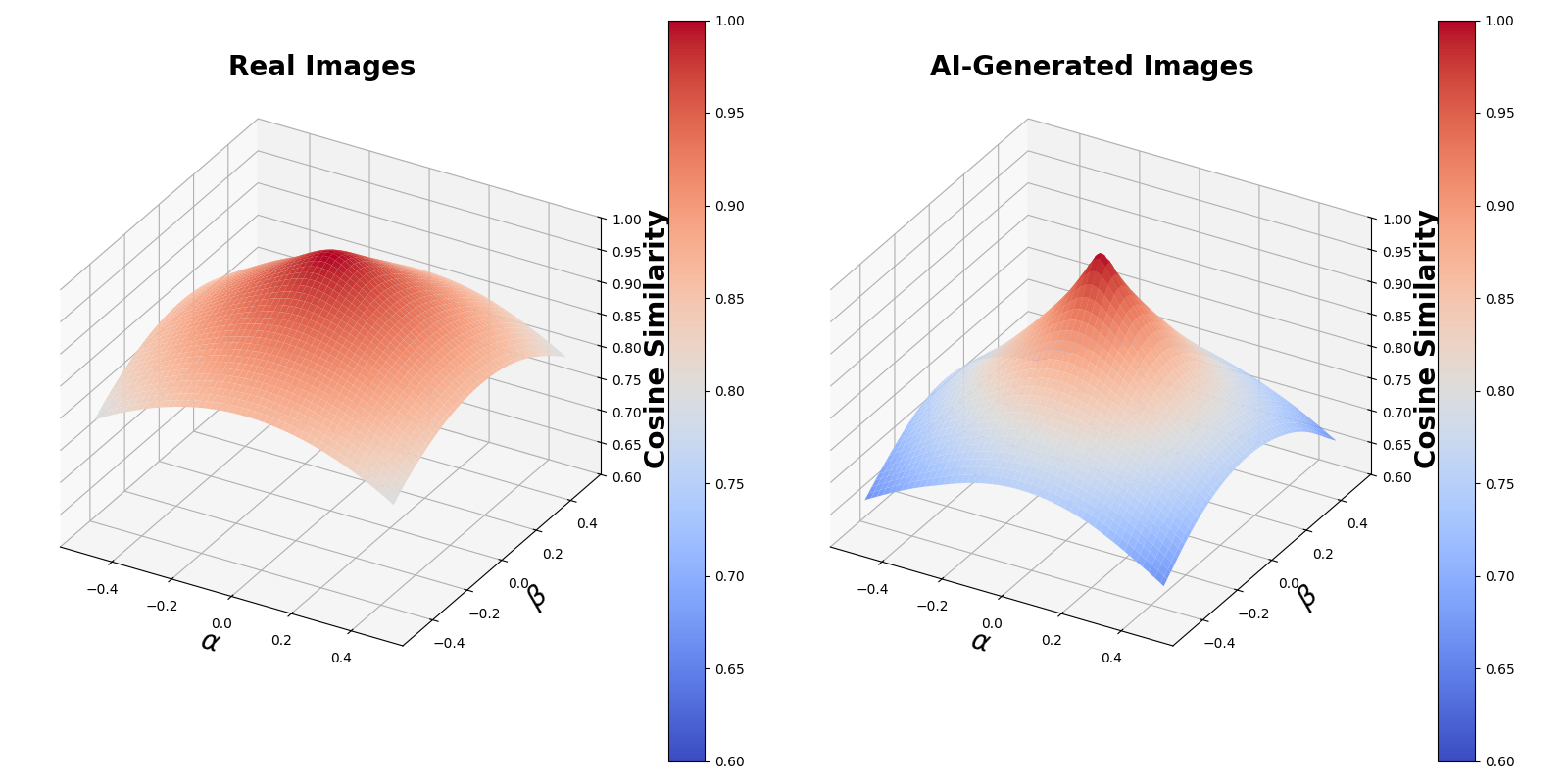}
    \caption{Loss landscape analysis for real and AI-generated images. We use the embedding of DINOV2 \cite{oquab2023dinov2} to compute the cosine similarity between an original image and its perturbed version by interpolating two random directions with additive Gaussian noise in pixel space, where the perturbation strengths are denoted by $\alpha$ and $\beta$. The results are averaged over 100 images. The real images are sampled from ImageNet, while the AI-generated ones are generated by the ablated diffusion mode (ADM) \cite{dhariwal2021diffusion}.
    The cosine similarity analysis shows that AI-generated images are more sensitive to Gaussian perturbations than real images.}
    \label{fig:real-fake}
\end{figure*}

\section{Model Output Safety: AI-Generated Content Detection}

With the rapidly increasing ability of GenAI tools to generate realistic, high-quality, and creative content across modalities, including but not limited to text, image, audio, and video, procedures to ensure that AI-generated content (AIGC) can be reliably validated are critical to the responsible use and governance of GenAI. Notable challenges associated with AIGC detection include misuse to create and disseminate disinformation at scale (e.g., AI-generated deepfakes), contamination of future training data, intellectual property protection, hallucinated responses, and inappropriate or unethical content generation. It is worth noting that watermarking is only a partial and somewhat limited solution to the challenge of AIGC detection, because watermarking only regulates responsible GenAI service providers, but not bad actors. A bad actor can train his own GenAI model or use any unregulated GenAI tool to create AIGC without watermarking. Therefore, watermarking only provides passive protection for AIGC, and reliable AIGC detectors without watermark assumption are in high demand as proactive AI safety tools. In this section, we discuss two AIGC detectors for image and text that are enhanced by signal processing techniques.

\paragraph*{AI-generated Image Detection via Sensitivity Analysis}
For the detection of AI-generated images, we focus on training-free approaches that are built upon off-the-shelf AI models,
as recent work has shown comparable or even better detection performance of training-free methods than training-based approaches \cite{he2024rigid,tsai2024understanding}.
Recall that sensitivity analysis computes a metric $M$ and uses it for hypothesis testing. Ricker et al. propose AEROBLADE \cite{ricker2024aeroblade}, a detector that uses the reconstruction error obtained from the pre-trained autoencoder of a latent diffusion model (LDM) for detection, where AI-generated images are found to have lower reconstruction errors than real images. He el al. propose RIGID \cite{he2024rigid}, which computes the cosine similarity between the representations of a pair of the original and perturbed images subject to Gaussian noise, using the DINOV2 model \cite{oquab2023dinov2}
for feature extraction. They find that real images tend to have higher cosine similarity to their perturbed versions than AI-generated images.

\begin{table}[t]
\caption{AUROC scores of training-free AI-generated image detectors on three datasets: ImageNet, LSUN-Bedroom, and DF40 (deepfakes).}
\begin{adjustbox}{max width=1\columnwidth}
\centering
\begin{tabular}{@{}l|lll@{}}
\toprule
Method   & ImageNet & LSUN-Bedroom & DF40 (Deepfake) \\ \midrule
AEROBLADE & 0.593    & 0.595        & 0.526           \\
RIGID     & 0.867    & 0.877        & 0.735           \\ \bottomrule
\end{tabular}
\end{adjustbox}
\label{table:image-detection}
\end{table}

Using the two-dimensional loss landscape analysis, we explore the landscape of cosine similarity $F(\alpha,\beta)$  between an original image and its perturbed version in the embedding space of DINOV2, where two independent noises drawn from the standard Gaussian distribution with their respective scaling factor $\alpha$ and $\beta$ are added to the pixel values of the original image. Figure \ref{fig:real-fake} shows the average landscape of cosine similarity over 100 real and AI-generated images, respectively, where the real ones are sampled from ImageNet, and the AI-generated ones are generated by the ablated diffusion mode (ADM) \cite{dhariwal2021diffusion}.
It can be observed that real images attain a higher similarity after perturbation than AI-generated images. The reason is that DINOV2 is trained only on real images and their data augmentations, which makes its embedding space less sensitive to noise perturbations on real images. Following \cite{he2024rigid}, we compare AEROBLADE and RIGID on three datasets consisting of real and AI-generated images: ImageNet, LSUN-Bedroom, and DF40 (deepfakes) \cite{yan2024df}. The AI-generated images are collected from a variety of generative models, including different variants of diffusion models, generative adversarial networks (GANs), and variational autoencoders (VAEs). Table \ref{table:image-detection} reports their AUROC scores, where higher values indicate better detection performance. RIGID is shown to outperform AEROBLADE. The difference can be explained by the fact that AEROBLADE relies on the pre-trained autoencoder of an LDM for detection, which may have limited generalization ability for images generated by different models. On the other hand, RIGID is agnostic to image generation models.

\paragraph*{AI-generated Text Detection via Adversarial Learning}

Due to the nature of autoregressive and probabilistic sampling of next-token prediction in LLMs, statistical approaches that derive statistics from LLMs for AI-generated text detection are popular baselines. Examples include log-likelihood (log p), token ranks (rank), and predictive entropy (entropy). Mitchell et al. propose DetectGPT \cite{mitchell2023detectgpt}, a detection method that adopts a log-likelihood ratio test between the original text and its perturbed-and-reworded versions.
However, recent studies have shown that many AI-generated text detectors are vulnerable to AI paraphrasing \cite{sadasivan2023can} -- using an LLM to rewrite the original AI-generated text can circumvent the reliability of many detectors. To address this challenge, Hu et al. \cite{hu2023radar} propose a detection method called RADAR, which uses adversarial learning to iteratively train a detector and a paraphraser to improve the robustness of AI-generated text detection.  During RADAR's training, the paraphraser uses proximal policy optimization to update its parameters, using the detector's predictions about its paraphrased text as a reward. The detector takes human-written text, original AI-generated text, and paraphrased AI-generated text as input, to maximize detection performance on human and AI-generated text. This iterative training process continues until detection performance is saturated on a validation dataset. With this adversarial learning setup, the detector learns to be robust in detecting both original and paraphrased AI-generated text.

Following the same experimental setup as in \cite{hu2023radar}, we collect human-written and AI-generated text from four different text datasets, including text summarization, question answering, Reddit's writing prompts, and TOFEL exam essays, where 8 different LLMs are used for text generation. In the evaluation, we also use OpenAI’s GPT-3.5-Turbo API as a paraphraser to test the robustness of detectors.  This paraphraser was not used to train any of the detectors. In addition to the above detectors, we include OpenAI's RoBERTa-based detector\footnote{https://huggingface.co/openai-community/roberta-base-openai-detector} for comparison. Table \ref{table:text-detection} compares the detection performance of different methods using the AUROC score. In the scenario of no paraphrasing, many detectors can perform well, achieving  AUROC scores as high as 0.904. However, when tested against paraphrasing, all detectors show a significant degradation in detection performance (up to more than 50\% decrease), except for RADAR. The stability of RADAR's performance can be attributed to its use of adversarial learning for robust detection.

\begin{table*}[t]
\caption{AUROC scores of AI-generated text detectors under two scenarios: \textit{without paraphrasing} and \textit{with paraphrasing} on the original AI-generated text. The paraphraser is OpenAI's GPT-3.5-Turbo API. Except for RADAR, all detectors show a performance drop when detecting paraphrased AI-generated text.}
\centering
\begin{tabular}{@{}l|lllllll@{}}
\toprule
Scenario             & log p & rank  & log rank & entropy & DetectGPT & OpenAI & RADAR \\ \midrule
Without Paraphrasing & 0.887 & 0.755 & 0.904    & 0.472   & 0.864     & 0.900  & 0.856 \\
With Paraphrasing    & 0.442 & 0.461 & 0.426    & 0.651   & 0.434     & 0.627  & 0.857 \\ \bottomrule
\end{tabular}
\label{table:text-detection}
\end{table*}

\section{Discussion: When AGI Means Artificial Good Intelligence}

This section discusses several insights to accelerate research and innovation in safety from a signal processing perspective. While the trends of GenAI lean toward the mindset of creating frontier AI technology by moving fast and scaling everything, especially for frequent model/product releases and intensified training and inference resources, embracing boldness and imperfection is becoming the new norm for AI safety. However, we argue that the development and use of scientific methods and tools, such as signal processing, is the most important milestone that must be achieved to claim an undisputed victory in GenAI. In particular, the computational AI safety framework introduced in this paper can be further developed into a new discipline that encompasses a comprehensive list of topics such as safety risk exploration (e.g., benchmarks, red-teaming, adversarial testing, and attacks), safety risk mitigation (e.g., detection, model updates, and safety-enhanced training), and evaluation (e.g., risk-capability analysis, safety certification, human-centered interaction, and governance). By studying and practicing computational AI safety, we believe we are just at the beginning of a new era and a bright future towards the ultimate goal of building a \textit{Safety Generalist} to ``use AI to govern AI,'' which involves autonomous exploration, mitigation, and evaluation of safety risks with human oversight as a controllable knob. The methods and techniques discussed in this paper only scratch the surface of signal processing for computational AI safety, and open science and community support are the organic engines to accelerate the momentum.

\paragraph*{\textbf{We don't (and shouldn't) need to build safety guardrails from scratch for GenAI}}
While GenAI may seem like a new technology, many lessons and past experiences learned from safeguarding pre-GenAI machine learning (ML) systems can be used as a foundation to quickly build safety guardrials for GenAI. This kind of ``knowledge transfer'' can avoid the mistakes of reinventing the wheel and effectively shorten the development of safety tests and patches for emerging GenAI systems. In addition, as the context and model size of GenAI continue to grow, building safety guardrails from scratch is not only time-consuming, but also resource-intensive. As an example, Table \ref{table:AdvML} shows a one-to-one mapping between the research topics in adversarial robustness \cite{chen2023book} for pre-GenAI machine learning systems and the seemingly new safety challenges in GenAI. The problems therein can be formulated using the same principles, and therefore the methods and tools developed for classical ML can be easily extended to study safety in GenAI. Specifically, adversarial examples stem from finding problematic data inputs with minimal and human-imperceptible changes to cause misclassification of the target ML system, while jailbreak features prompt modification and stealth to induce harmful output for GenAI. Data poisoning and backdoor are training-phase risks that are directly related to biased or malicious instruction tuning and model fine-tuning for GenAI. Out-of-distribution generalization in ML explores the robustness under semantically similar but unseen changes during training, while alignment in GenAI means teaching the model to learn the values and principles implicit in the alignment data and generalize beyond the provided context. Finally, model reprogramming \cite{chen2024model} explores cross-domain ML by attaching input and output transformation functions to a weight-frozen ML model, similar to prompt injection in GenAI, which manipulates a target GenAI system to produce an incompliant output via prompt hacking.

\begin{table*}[t]
\caption{The mapping between the topics of adversarial robustness in classical (pre-GenAI) machine learning systems and the seemingly new safety challenges in GenAI. Many techniques developed for classical machine learning can be easily extended to study the associated safety challenges in GenAI.}
\centering
\begin{tabular}{@{}l|l@{}}
\toprule
Classical Machine Learning       & GenAI                            \\ \midrule
Adversarial Example                & Jailbreak Prompt                 \\
Data Poisoning / Backdoor          & Biased or Malicious Instructions \\
Out-of-Distribution Generalization & Alignment                        \\
Model Reprogramming                & Prompt Injection                 \\
\bottomrule
\end{tabular}
\label{table:AdvML}
\end{table*}

\paragraph*{\textbf{Open Challenges and Opportunities}} To strengthen the link between signal processing and AI safety, we highlight some open research challenges and opportunities.
\begin{itemize}[leftmargin=*]
    \item Computational safety for multi-modal GenAI: GenAI technology is on its way to accepting inputs from multiple data types and potentially generating outputs with mixed modalities. While multi-modal GenAI may introduce increased safety risks by providing additional attack surfaces, techniques such as heterogeneous (multi-view) signal processing applied to multiple data sources can be explored for safety research.
    \item Computational safety for agentic and physical GenAI: The next generation of GenAI is to be empowered with the ability to act on behalf of the user by autonomously creating agentic workflows and potentially participating in multi-agent interactions. In addition, GenAI is expected to thrive in embodied agentic systems (e.g., AI humanoid robots), extending the discussion of AI safety from the digital space to the physical world \cite{tang2024defining}. Control-based signal processing methods may be a promising direction to explore agentic and physical safety for GenAI.     
    \item Practical AI safety guardrails: In addition to addressing the inherent trade-offs between safety and capability, the practical implementation of AI safety guardrails must be cost-effective in terms of computation, memory, and agility. Compression and high-dimensional statistics in signal processing can be powerful tools to address long-context and real-time requirements. 
    \item Understanding the fundamental limits of AI safety:  To better understand and design alignment training and reward modeling in GenAI, theoretical signal processing approaches can be used to characterize the information-theoretic limits, statistical properties, or computational optimality for GenAI. Furthermore, model-based signal processing can be explored as a potential approach to overcome the limitations of GenAI. 
\end{itemize}

\paragraph*{\textbf{May AGI Mean ``Artificial Good Intelligence''}}
Although AGI is often used as shorthand for Artificial General Intelligence, and there are many expectations that AGI may be a near-term possibility, we argue that computational safety is a critical prerequisite for the development and deployment of safe and responsible AI technology. Through the presented computational AI safety framework,
  this paper's primary objective is to elucidate the pivotal role of signal processing in the examination and resolution of numerous foundational and pragmatic safety concerns pertaining to GenAI. We posit that this paper can serve as a catalyst for researchers and practitioners to accord greater consideration to signal processing techniques as a means of ``good intelligence.'' Should AGI remain the overarching objective, it is of the utmost importance to ensure that this term represents ``Artificial Good Intelligence.'' Moreover, while safety is becoming the new arms race among leading GenAI stakeholders, we strongly believe that \textit{safety should not be a competition and a race to the bottom}. Instead of a winner-take-all game, AI safety should be considered a public good through open science and borderless collaboration. After all, the true worth of AI safety is as useful and meaningful as the individuals who can benefit from it.

\section{Conclusion}
This paper presents a computational safety framework for GenAI through the lens of signal processing. We show that many problems that arise in AI safety can be formulated in a unified hypothesis testing scheme. In particular, we study two representative use cases in AI safety, jailbreaks and AI-generated content, and elucidate how signal processing techniques can be used to build reliable detectors and develop efficient mitigation strategies. Our results suggest the effectiveness of hypothesis testing as a fundamental methodology for improving AI safety.
Finally, we articulate our views on the sustainability and growth of AI safety research and technology adoption, and why signal processing, a foundational technique for pattern recognition and machine intelligence, is a promising approach to accelerating its progress.

\section*{Acknowledgment}
The authors thank Mr. Xiaomeng Hu and Mr. Zhiyuan He for their help in providing the materials for the figures
and numerical results.

\bibliography{adversarial_learning}
\bibliographystyle{icml2026}



\end{document}